\documentclass[runningheads]{llncs}
\usepackage[T1]{fontenc}
\usepackage{graphicx}
\usepackage{booktabs}
\usepackage{accvabbrv}
\usepackage{amsmath}
\usepackage{cite}
\usepackage{hyperref}
\usepackage[accsupp]{axessibility}

\usepackage{color}

\begin{document}
\title{TANet: Triplet Attention Network for All-In-One Adverse Weather Image Restoration}
\titlerunning{TANet}
%
\author{Hsing-Hua Wang\inst{*, 1} \and
Fu-Jen Tsai\inst{*, 1} \and
Yen-Yu Lin\inst{2} \and
Chia-Wen Lin\inst{1}}
\authorrunning{H.-H. Wang et al.}
%
\institute{National Tsing Hua University, Taiwan \\ 
\email{xhwchris@gapp.nthu.edu.tw, fjtsai@gapp.nthu.edu.tw, cwlin@ee.nthu.edu.tw} \and
National Yang Ming Chiao Tung University, Taiwan \\
\email{lin@cs.nycu.edu.tw}
}
\maketitle              
\begingroup
\renewcommand\thefootnote{}\footnotetext{* equal contribution}
\addtocounter{footnote}{-1}
\endgroup

\begin{abstract}
Adverse weather image restoration aims to remove unwanted degraded artifacts, such as haze, rain, and snow, caused by adverse weather conditions. Existing methods achieve remarkable results for addressing single-weather conditions. However, they face challenges when encountering unpredictable weather conditions, which often happen in real-world scenarios. Although different weather conditions exhibit different degradation patterns, they share common characteristics that are highly related and complementary, such as occlusions caused by degradation patterns, color distortion, and contrast attenuation due to the scattering of atmospheric particles. Therefore, we focus on leveraging common knowledge across multiple weather conditions to restore images in a unified manner. 
In this paper, we propose a Triplet Attention Network (TANet) to efficiently and effectively address all-in-one adverse weather image restoration. TANet consists of Triplet Attention Block (TAB) that incorporates three types of attention mechanisms: Local Pixel-wise Attention (LPA) and Global Strip-wise Attention (GSA) to address occlusions caused by non-uniform degradation patterns, and Global Distribution Attention (GDA) to address color distortion and contrast attenuation caused by atmospheric phenomena. By leveraging common knowledge shared across different weather conditions, TANet successfully addresses multiple weather conditions in a unified manner. Experimental results show that TANet efficiently and effectively achieves state-of-the-art performance in all-in-one adverse weather image restoration. The source code is available at \href{https://github.com/xhuachris/TANet-ACCV-2024}{https://github.com/xhuachris/TANet-ACCV-2024}.    

\end{abstract}
\section{Introduction}

\begin{figure}[t]
\begin{center}
\includegraphics[width=1\columnwidth]{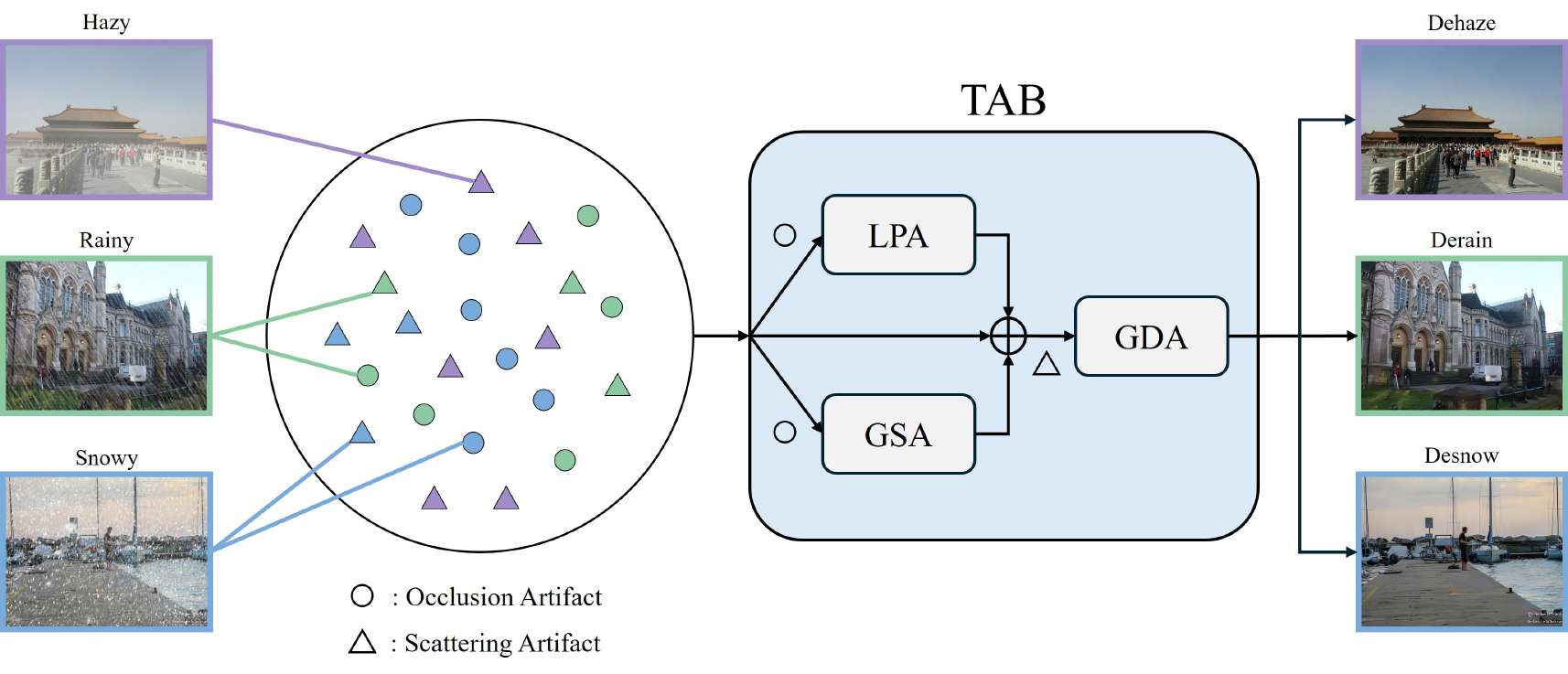}
\end{center}
\caption{In TANet, we utilize Triple Attention Block (TAB) to effectively address occlusion and scattering artifacts caused by adverse weather conditions. In TAB, we utilize a Local Pixel-wise Attention (LPA) and a Global Strip-wise Attention (GSA) to address non-uniform degradation patterns. In addition, we utilize Global Distribution Attention to handle unwanted scattering artifacts caused by atmospheric phenomena. 
}
\label{fig:intro}
\end{figure}

Adverse weather conditions, such as haze, rain, and snow, often cause unwanted artifacts that degrade the visual quality of images. These degradation patterns obscure the structure and details of images, severely affecting many downstream vision tasks. Adverse weather image restoration aims to remove undesirable degradation patterns from a single degraded image, which is a highly ill-posed problem as weather conditions are typically non-uniform and time-varying, making adverse weather image restoration a challenging task that has been widely studied in computer vision~\cite{Dark_Channel_Prior, Kang_2012_TIP, Ma_2022_CVPR}.  

Adverse weather image restoration has achieved remarkable progress with the development of deep learning. Several studies have reached promising results on single-weather image restoration task, including dehazing~\cite{Wu_2021_CVPR,guo2022dehamer,song2023vision,9010659,Hardgan,FSDGN,qin2020ffa}, deraining~\cite{Jiang_2022_ACMMM,Hu_2019_CVPR,Wang_2020_CVPR,Wang_2019_CVPR,8099669,Kui_2020_CVPR,Li_2019_CVPR}, and desnowing~\cite{JSTASRChen,chen2021all,8291596,zhang2021deep,7934435}.
Although previous works significantly enhance visual quality under specific weather conditions, the requirement for prior knowledge of specific weather conditions limits their applicability in unpredictable real-world scenarios. Since weather conditions can change over time, addressing unpredictable weather conditions becomes more challenging for image restoration models. Therefore, there is a need to develop an all-in-one image restoration network that can handle multiple weather conditions in a unified manner without relying on weather-specific prior knowledge.      

Recently, several works~\cite{AirNet,Valanarasu_2022_CVPR,Chen2022MultiWeatherRemoval,Park_2023_CVPR,Patil_2023_ICCV,potlapalli2023promptir}  have focused on addressing multiple adverse weather conditions in a unified manner. To handle various degradation patterns, some methods resort to extracting weather-specific features through the design of weather-specific components, such as weather type queries~\cite{Valanarasu_2022_CVPR} and degradation-conditioned prompts~\cite{potlapalli2023promptir}. 
However, despite different types of weather conditions causing different degradation patterns, these degradation patterns are highly related and complementary. For example, the rain and snow masks often exhibit occluded artifacts with various directions and magnitudes. The scattering of atmospheric particles often causes color distortion and contrast attenuation under adverse weather conditions~\cite{8259491}. Therefore, instead of designing weather-specific modules to distinguish different types of weather conditions, we aim to leverage common knowledge across degradation patterns for addressing all-in-one adverse weather image restoration. This approach allows us to build an effective and efficient image restoration network based on the inductive bias of adverse weather conditions.    

In this paper, we propose a Triplet Attention Network (TANet) which consists of a Triplet Attention Block (TAB) with three types of attention modules to effectively and efficiently address all-in-one adverse weather image restoration, including dehazing, deraining, and desnowing. 
As shown in Figure~\ref{fig:intro}, hazy, rainy, and snowy patterns contain occluded artifacts that are typically non-uniform and vary in size. To effectively handle these degraded patterns, TAB incorporates two types of spatial attention modules. Specifically, TAB utilizes local pixel-wise attention (LPA) to capture local spatial information and global strip-wise attention (GSA), including horizontal and vertical strip attention, to capture global spatial information, which allows TAB to handle rainy and snowy patterns with various orientations and magnitudes. Additionally, the multi-scale design also enables TAB to effectively address non-uniform degraded patterns under various adverse weather conditions.   

Moreover, images taken under adverse weather conditions often suffer from color distortion and contrast attenuation due to the scattering of atmospheric particles. The distribution of atmospheric particles highly affects the intensity of scattering. Therefore, TAB incorporates global distribution attention (GDA) to capture the varying distribution of atmospheric particles. Specifically, since the distribution of atmospheric particles varies among degraded images, TAB utilizes instance normalization to perform feature normalization within each image, allowing adaptively adjustment of the feature distribution in degraded images. As a result, by using three attention mechanisms, TAB can focus on local, global, and distribution information, leading to an efficient and effective network by leveraging the inductive bias of adverse weather conditions. Experimental results demonstrate that TANet achieves state-of-the-art results on both synthetic and real-world adverse weather image restoration datasets. The contributions of TANet can be summarized as follows: 
\begin{itemize}
    \item We propose a Triplet Attention Network (TANet), an efficient and effective all-in-one image restoration network for addressing adverse weather conditions. 
    \item TANet utilizes Triplet Attention Block (TAB) that leverages the inductive bias of adverse weather conditions to simultaneously capture local, global, and distribution information to remove degraded patterns.
    \item Experimental results demonstrate that TANet achieves state-of-the-art results on both synthetic and real-world adverse weather image restoration datasets. 
    
\end{itemize}

\section{Related Work}
\subsubsection{Single Degradation Image Restoration}
With the development of deep learning, single-weather image restoration has reached remarkable results, including image dehazing~\cite{ Wu_2021_CVPR,Zheng_2021_CVPR,guo2022dehamer,song2023vision,9010659,Hardgan,FSDGN,qin2020ffa,Qiu_2023_ICCV}, deraining~\cite{Chen_2021_CVPR,Hu_2019_CVPR,Wang_2020_CVPR,Wang_2019_CVPR,8099669,Li_2019_CVPR,li2018recurrent,Kui_2020_CVPR,9096546,Chen_2023_CVPR,xiao2022image,Zamir2021Restormer}, and desnowing~\cite{8291596,zhang2021deep,JSTASRChen,chen2021all,7934435}. For image dehazing, several studies improved dehazing performance by extracting haze-related features. Deng~\etal~\cite{Hardgan} proposed a haze-aware representation distillation module to extract haze-aware features. Guo~\etal~\cite{guo2022dehamer} proposed a CNN and Transformer hybrid network with transmission-aware position embedding to address hazy patterns. 
For image deraining, several studies~\cite{li2018recurrent,Kui_2020_CVPR,9096546} utilized recurrent-based networks to progressively remove rain streaks. Li~\etal~\cite{li2018recurrent} proposed a recurrent network that utilized dilated convolutions to enhance receptive fields. Jiang~\etal~\cite{Kui_2020_CVPR} proposed a multi-scale pyramid architecture to recurrently remove rain streaks in a coarse-to-fine manner. 
For image desnowing, previous studies mainly focus on addressing snowy patterns with various sizes. Chen~\etal~\cite{JSTASRChen} proposed a size-aware and transparency-aware network for removing snow and veil effects. Zhang~\etal~\cite{zhang2021deep} proposed a multi-scale snow removal network that utilized semantic and geometric guidance in a coarse-to-fine manner. Although these weather-specific methods achieve promising results for specific weather conditions, their extensibility to other weather conditions remains a concern due to the design of weather-specific architectures. Consequently, several works have proposed generic image restoration networks to address multiple degraded patterns.

\subsubsection{Multiple Degradation Image Restoration}
Instead of designing a degradation-specific architecture, some methods~\cite{Zamir2021MPRNet,chen2022simple,cui2023focal} develop a generic model that can be trained on different tasks to handle various types of degradation. Zamir~\etal~\cite{Zamir2021MPRNet} propose a multi-patch architecture to recurrently restore degraded images in a cross-to-fine manner. Chen~\etal~\cite{chen2022simple} design a nonlinear activation-free network for generic image restoration without using nonlinear activation functions. Cui~\etal~\cite{cui2023focal} propose a dual-domain selection network that contains spatial and frequency selection to extract crucial features for image restoration. 
Although these methods use generic architectures to address various types of degradation, they require manual switching of pre-trained models for addressing different types of degradations. This is not suitable for real-world applications, where degraded patterns are typically unpredictable. Furthermore, although it is possible to optimize these generic image restoration methods in an all-in-one manner, they often ignore the inductive bias of adverse weather conditions, including occluded artifacts caused by degraded patterns, color distortion, and contrast attenuation caused by scattering of atmospheric particles, constraining their performance when addressing adverse weather conditions. Therefore, we propose TANet that leverages the inductive bias of adverse weather conditions to address various unknown weather conditions in an all-in-one manner.     
\subsubsection{All-in-one Image Restoration}
Compared to single and multi degradation image restoration, all-in-one image restoration~\cite{AirNet,Valanarasu_2022_CVPR,Chen2022MultiWeatherRemoval,Park_2023_CVPR,Patil_2023_ICCV,potlapalli2023promptir} aims to address various types of degradations in a unified model, which offers several advantages, such as strong generalization ability and reduced storage requirements.    
To address various weather conditions in an all-in-one manner, several studies have incorporated weather-specific modules into all-in-one restoration networks for adaptively addressing various unknown degradations. Chen~\etal~\cite{Chen2022MultiWeatherRemoval} proposed to use a teacher-student architecture by distilling knowledge from multiple weather-specific teacher models. However, the need to learn multiple weather-specific teacher models significantly increases the computational costs during training. Valanarasu~\etal~\cite{Valanarasu_2022_CVPR} proposed to utilize learnable weather types queries in Transformer, and Potlapalli~\etal~\cite{potlapalli2023promptir} proposed to incorporate learnable prompts regarding degradation information in Transformer to achieve all-in-one image restoration. However, aside from learning additional weather-specific parameters for addressing all-in-one image restoration, we note that different types of weather conditions are highly related and complementary. These degraded patterns share common characteristics, such as occlusions caused by degraded patterns, color distortion and contrast attenuation due to the scattering of atmospheric particles.  
Therefore, TANet utilizes triplet attention models to simultaneously address occlusion and scattering problems in a unified manner, which leverages the inductive bias of adverse weather conditions and effectively and efficiently addresses adverse weather image restoration in an all-in-one manner.

\section{Proposed Method}

\subsection{Overview}

\begin{figure}[t]
\begin{center}
\includegraphics[width=1\columnwidth]{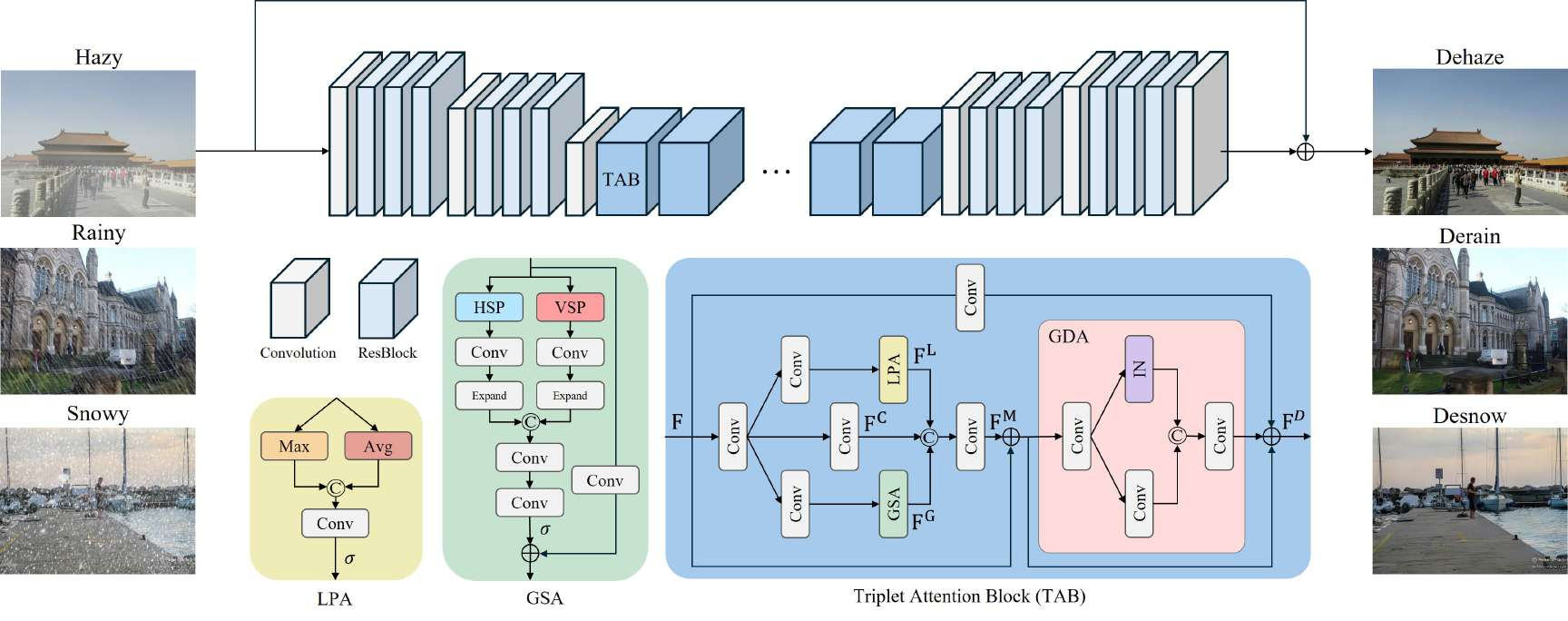}
\end{center}
\caption{Architecture of TANet. TANet is an encoder-decoder network comprising several Triplet Attention Blocks (TAB). In TAB, we utilize Local Pixel-wise Attention (LPA), Glbal Strip-wise Attention (GSA), and Global Distribution Attention (GDA) to effectively degradation patterns with occlusion and scattering artifacts. $\textcircled{c}$ and $\oplus$ denote concatenation and addition.
}
\label{fig:architecture}
\end{figure}

In this paper, we propose a Triplet Attention Network (TANet) to address adverse weather image restoration in an all-in-one manner. Unlike previous methods~\cite{AirNet,Valanarasu_2022_CVPR,Chen2022MultiWeatherRemoval,Park_2023_CVPR,Patil_2023_ICCV,potlapalli2023promptir} that extract weather-specific features, TANet leverages generic knowledge across various degradations types based on the inductive bias of adverse weather conditions.
As shown in Figure~\ref{fig:architecture}, TANet is a CNN-based encoder-decoder network that starts with two Feature Embedding Layers (FEL) to downscale features, each comprising a convolutional layer with three residual blocks. Next, we stack several Triplet Attention Blocks (TAB). Each TAB comprises Local Pixel-wise Attention (LPA), Global Strip-wise Attention (GSA), and Global Distribution Attention (GDA) to address occlusion and scattering artifacts simultaneously. Lastly, we utilize another two FELs to upscale the attended features for reconstructing a clean image. Subsequent sections will elaborate on TAB components, including LPA, GSA, GDA, and the final loss function used for optimizing TANet.

\subsection{Triplet Attention Block (TAB)}
As shown in Figure~\ref{fig:architecture}, TAB comprises Local Pixel-wise Attention (LPA), Global Strip-wise Attention (GSA), and Global Distribution Attention (GDA) to address occlusion and scattering artifacts simultaneously. 
Let input features be ${F} \in \mathbb{R}^{H \times W \times C}$, where $H$, $W$, and $C$ denote height, width, and the number of channels, respectively. We process $F$ through a convolutional layer followed by three parallel branches to generate multi-scale features $F^{L}$, $F^{G}$, and $F^{C} \in \mathbb{R}^{H \times W \times C}$, where $F^{L}$ and $F^{G}$ are the outputs of 
LPA and GSA as 
\begin{equation}
\begin{gathered}
    F^{L} = \mathrm{LPA}(\mathrm{Conv}(\mathrm{Conv}(F))), \\ 
    F^{G} = \mathrm{GSA}(\mathrm{Conv}(\mathrm{Conv}(F))), \\
    F^{C} = \mathrm{Conv}(\mathrm{Conv}(\mathrm{Conv}(F))),
\end{gathered}
\end{equation}
Next, we concatenate them followed by a convolutional layer with a residual connection to generate multi-scale attended features $F^M\in \mathbb{R}^{H \times W \times C}$ as    
\begin{equation}
    F^{M} = \mathrm{Conv}(\mathrm{Concate}(F^{L},F^{G},F^{C})) + F,
\end{equation}
In this step, we fuse multi-scale features to handle non-uniform degradation patterns under adverse weather conditions.
After addressing degradation patterns that obscure object structure, we utilize GDA to address color distortion and contrast attenuation caused by the scattering of atmospheric particles. Therefore, we process $F^{M}$ through GDA with two residual connections for globally addressing the unknown distribution of atmospheric particles as  
\begin{equation}
    F^{D} = \mathrm{GDA}(F^M) + \mathrm{Conv}(F) + F^{M},
\end{equation}
where $F^{D}\in \mathbb{R}^{H \times W \times C}$. Following, we describe LPA, GSA, and GDA in detail.

\subsubsection{Local Pixel-wise Attention (LPA)}
As shown in Figure~\ref{fig:architecture}, to effectively address occlusions caused by short-range degradation patterns, TAB utilizes local pixel-wise attention (LPA) to extract local spatial features.
Motivated by~\cite{CBAM}, let input features be ${F} \in \mathbb{R}^{H \times W \times C}$, we utilize average pooling and max pooling operations along the channel axis to generate two types of feature maps: $F^{avg}\in \mathbb{R}^{H \times W \times 1}$ and $F^{max}\in \mathbb{R}^{H \times W \times 1}$.
Next, we concatenate them followed by a convolutional layer with a sigmoid function to generate the spatial attention feature $F^{L} \in \mathbb{R}^{H \times W \times C}$ as 
\begin{equation}
\begin{gathered}
    F^{L} = \sigma({\mathrm{Conv}(\mathrm{Concate}(AvgPool(F),MaxPool(F))})),
    \label{}
\end{gathered}
\end{equation}
where $\sigma$ denotes the sigmoid function. 

\subsubsection{Global Strip-wise Attention (GSA)}

\begin{figure}[t]
\begin{center}
\includegraphics[width=1\columnwidth]{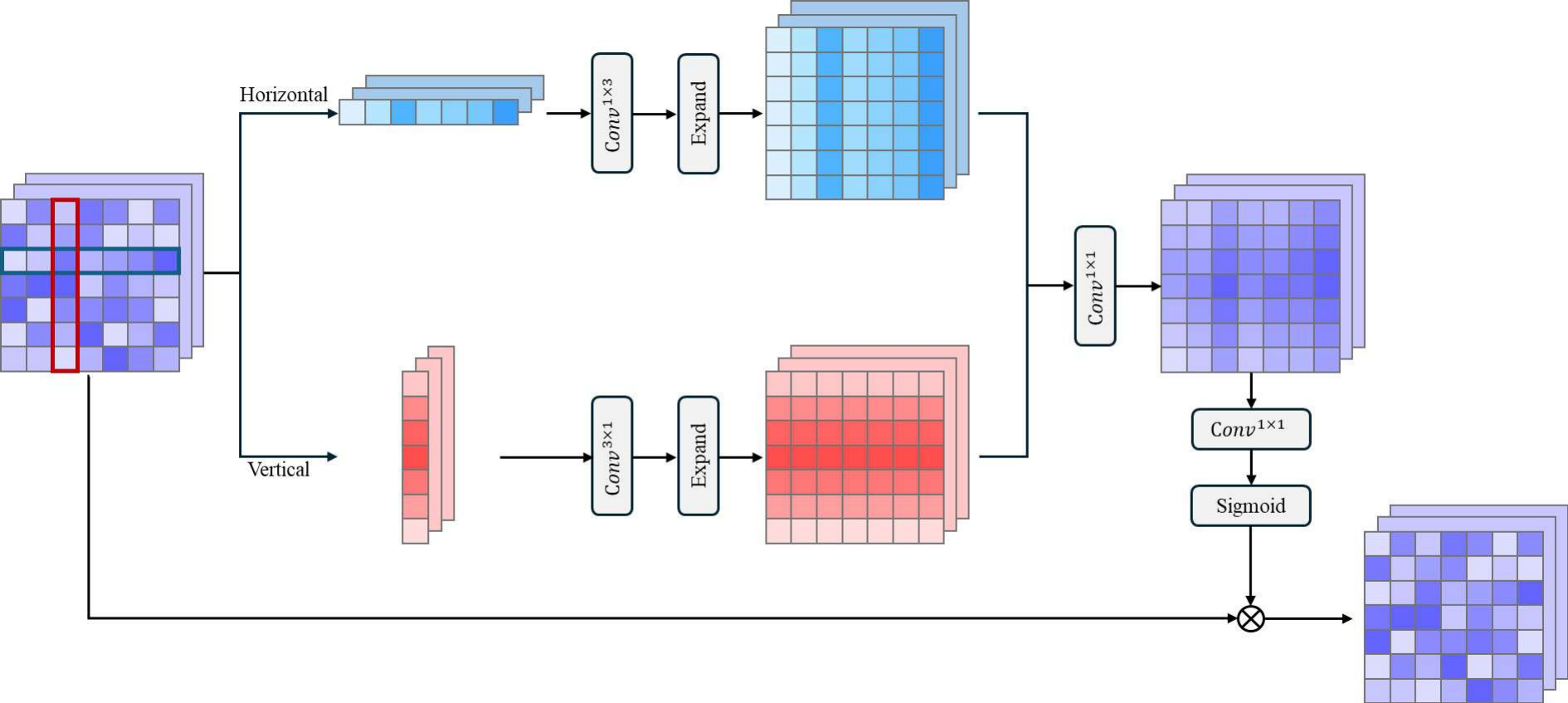}
\end{center}
\caption{Architecture of Global Strip-wise Attention (GSA). GSA utilizes horizontal and vertical strip pooling to project features in horizontal and vertical directions. After fusing horizontal and vertical attended features, GSA efficiently addresses degradation patterns with various orientations.
}
\label{fig:GSA}
\end{figure}

As shown in Figure~\ref{fig:GSA}, to effectively address occlusions caused by long-range degradation patterns, TAB utilizes GSA to extract global spatial features. Motivated by~\cite{hou2020strip}, since occluded artifacts, such as rainy and snowy patterns, contain degradation patterns with various orientations, we utilize strip-pooling that contains horizontal and vertical strip-shape pooling operations to project features into horizontal and vertical directions. This enables us to efficiently address long-range degradation patterns with various orientations.      
Let input features be ${F} \in \mathbb{R}^{H \times W \times C}$, we project $F$ into horizontal ${F^h} \in \mathbb{R}^{1 \times W \times C}$ and vertical ${F^v} \in \mathbb{R}^{H \times 1 \times C}$ features by long-range average pooling operation as 
\begin{equation}
\begin{gathered}
    F^{h}_{j,c} = \frac{1}{H}\sum_{0\leq i<H}{F_{i,j,c}}, \\
    F^{v}_{i,c} = \frac{1}{W}\sum_{0\leq j<W}{F_{i,j,c}}, 
    \label{}
\end{gathered}
\end{equation}
where $i$, $j$, and $c$ denote the index of height, width, and channel dimensions. 
Next, we use $1\times3$ and $3\times1$ convolutional layers to fuse horizontal $F^h$ and vertical $F^v$ features and expand their size to generate attended features $\Tilde{F}^{h}$ and $\Tilde{F}^{v} \in \mathbb{R}^{H \times W \times C}$  as 

\begin{equation}
\begin{gathered}
    \Tilde{F}^{h}_{i,j,c} = \mathrm{Expand}(\mathrm{Conv}^{1\times3}(F^{h}_{j,c})), \\
    \Tilde{F}^{v}_{i,j,c} = \mathrm{Expand}(\mathrm{Conv}^{3\times1}(F^{v}_{i,c})),
    \label{}
\end{gathered}
\end{equation}
where we use a copy operation for expansion. Lastly, we fuse two attended features by addition followed by a convolutional layer with a sigmoid function and multiply it with the original tensor as
\begin{equation}
    F^{G} = \sigma(\mathrm{Conv}(\Tilde{F}^{h} + \Tilde{F}^{v})) \otimes F, 
    \label{}
\end{equation}
where $F^G\in \mathbb{R}^{H \times W \times C}$ is the final output of GSA, and $\sigma$ and $\otimes$ denote the sigmoid function and element-wise multiplication.

\subsubsection{Global Distribution Attention (GDA)} 
Finally, as shown in Figure~\ref{fig:architecture}, to address the color distortion and contrast attenuation caused by the scattering of atmospheric particles, TAB utilizes GDA to capture the distribution of atmospheric particles. To adaptively capture the feature distribution of various degraded images, we adopt instance normalization that performs normalization within each instance. Motivated by~\cite{HINet},  we first split the input tensor ${F} \in \mathbb{R}^{H \times W \times C}$ along the channel dimension by a convolutional layer as 

\begin{equation}
    ({F_{1}}, {F_{2}}) = \mathrm{Split}({\mathrm{Conv}}(F)), 
    \label{}
\end{equation}
where $F_{1}$ and $F_{2}\in \mathbb{R}^{H \times W \times \frac{C}{2}}$.

Next, we process $F_1$ through an instance normalization (IN) layer while processing $F_2$ through a convolutional layer without using the instance normalization layer. This enables TAB to adaptively adjust the feature distribution while simultaneously preserving original information as    
\begin{equation}
\begin{gathered}
    F^{D} = \mathrm{Conv}(\mathrm{Concate}(\mathrm{IN}({F_{1}}),\mathrm{Conv}({F_{2}}))) + F, \\
    IN = \gamma(\frac{F_1-\mu(F_1)}{\sigma(F_1)}) + \beta
    \label{}
\end{gathered}
\end{equation}
where $F^D\in \mathbb{R}^{H \times W \times C}$ is the final output. $\mu$ and $\sigma$ denote the mean and variance operation. $\gamma$ and $\beta\in \mathbb{R}^{1 \times 1 \times \frac{C}{2}}$  are learnable affine parameters. 

\subsubsection{Loss Function}

To optimize TANet, we adopt the Charbonnier loss $\mathcal{L}_\mathrm{char}$ as  
\begin{equation}
\begin{gathered}
    \mathcal{L}_\mathrm{char}= \sqrt{||O-G||_2 + \epsilon^{2}}
    \label{}
\end{gathered}
\end{equation}
where $O$ and $G$ denote the restored image and the ground-truth image, and $\epsilon={10}^{-3}$. 
In addition, we adopt the FFT loss $\mathcal{L}_\mathrm{FFT}$ to supervise restoring images in the frequency domain as

\begin{equation}
\begin{gathered}
    \mathcal{L}_\mathrm{FFT} = ||\mathcal{F}(O)-\mathcal{F}(G)||_1,
    \label{}
\end{gathered}
\end{equation}
where $\mathcal{F}$ denotes the fast Fourier transform. Last, we optimize TANet by the total loss $\mathcal{L}_\mathrm{total}$ as
\begin{equation}
\begin{gathered}
    \mathcal{L}_\mathrm{total} = \mathcal{L}_\mathrm{char} + {\lambda}{\mathcal{L}_\mathrm{FFT}}
    \label{}
\end{gathered}
\end{equation}
where we experimentally set ${\lambda}$ to $1 \times 10^{-2}$.

\section{Experiments}

\begin{table*}[t!]
\small
\centering
\setlength{\tabcolsep}{1.5mm}
\caption{Quantitative comparisons on synthetic datasets, including  SOTS~\cite{Resides} for dehazing, Rain1400~\cite{8099669} for deraining, and Snow100K-L~\cite{8291596} for desnowing. We highlight the best two scored in bold and underline. The inference time is measured using images of size $256\times256$.}
\begin{tabular}{l|cccccc}
\noalign{\hrule height 1.0pt}
Model & Haze  & Rain & Snow & Average & Params (M) & Time (ms)     \\
\noalign{\hrule height 1.0pt} 
Transweather~\cite{Valanarasu_2022_CVPR}  & 27.66 & 29.14 & 26.17 & 27.66 & 38 & \underline{17} \\ 
KCKE~\cite{Chen2022MultiWeatherRemoval}  & 29.16 & 30.82 & 27.87 & 29.28 & 29 & 29 \\ 
WGWS~\cite{zhu2023Weather} & 26.64 & \underline{31.84} & 29.43 & 29.30 & 6 & 21 \\ 
PromptIR~\cite{potlapalli2023promptir} & 30.61 & 31.38 & 28.83 & 30.27 & 33 & 110 \\ 
NAFNet~\cite{chen2022simple} & 31.40 & 31.18 & 29.26 & 30.61 & 17 & 27  \\ 
FocalNet~\cite{cui2023focal} & 31.77 & 31.47 & 29.74 & 30.99 & \underline{4} & \bf12 \\ 
GRL~\cite{li2023grl} & 28.90 & 30.6 & 28.34 & 29.28 & \bf3 & 159 \\
MPRNet~\cite{Zamir2021MPRNet} & \underline{31.79} & 31.72 & \underline{30.05} & \underline{31.19} & 20 & 77 \\

TANet (Ours)  & \bf34.80 & \bf31.87 & \bf30.67 & \bf32.45 & 9 & 18 \\ 

\noalign{\hrule height 1.0pt}
\end{tabular}
\label{Tab:Test_Synthetic}
\end{table*}

\begin{table*}[t!]
\small
\centering
\setlength{\tabcolsep}{1.5mm}
\caption{Quantitative comparisons on the real-world WeatherStream~\cite{zhang2023weatherstream} dataset, including. We highlight the best two scored in bold and underline. The inference time is measured using images of size $256\times256$.}
\begin{tabular}{l|cccccc}
\noalign{\hrule height 1.0pt}
Model & Haze  & Rain & Snow & Average & Params (M) & Time (ms)     \\
\noalign{\hrule height 1.0pt} 
Transweather~\cite{Valanarasu_2022_CVPR}  & 19.03 & 22.19 & 20.77 & 20.66 & 38 & \underline{17} \\ 
KCKE~\cite{Chen2022MultiWeatherRemoval}  & 19.09 & \bf22.76 & 20.98 & 20.94 & 29 & 29 \\ 
WGWS~\cite{zhu2023Weather} & 18.82 & \underline{22.52} & 20.48 & 20.61 & 6 & 21 \\ 
PromptIR~\cite{potlapalli2023promptir} & \underline{20.76} & 22.17 & \underline{21.20} & \underline{21.38} & 33 & 110 \\ 
NAFNet~\cite{chen2022simple} & 20.07 & 22.17 & 20.76 & 21.00 & 17 & 27  \\ 
FocalNet~\cite{cui2023focal} & 19.25 & 22.34 & 21.06 & 20.88 & \underline{4} & \bf12 \\ 
GRL~\cite{li2023grl} & 19.15 & 21.91 & 20.69 & 20.58 & \bf3 & 159 \\
MPRNet~\cite{Zamir2021MPRNet} & 19.27 & 21.84 & 20.98 & 20.70 & 20 & 77 \\

TANet (Ours)  & \bf20.84 & 22.18 & \bf21.49 & \bf21.50 & 9 & 18 \\ 

\noalign{\hrule height 1.0pt}
\end{tabular}
\label{Tab:Test_WeatherStream}
\end{table*}

\subsection{Datasets and Implementation Details}
\subsubsection{Datasets.}\label{Datasets}
We utilize three synthetic datasets to optimize TANet, including RESIDE~\cite{Resides} for dehazing, Rain1400~\cite{8099669} for deraining, and Snow100K~\cite{8291596} for desnowing. Specifically, RESIDE consists of the ITS dataset containing 110,000 indoor training pairs, and the OTS dataset containing 313,950 outdoor training pairs. Rain1400 has 12,600 training pairs, and Snow100K has 100,000 training pairs. We follow~\cite{zhu2023Weather,Chen2022MultiWeatherRemoval} to uniformly sample 5,000 training pairs from each dataset and mix them to form a training set of 15,000 training pairs. For fair comparisons, we optimize TANet and all compared methods on this mixed training set. 
For evaluation, we utilize three synthetic and one real-world testing sets to demonstrate the effectiveness of TANet. For synthetic testing sets, we select SOTS~\cite{Resides} testing set that contains 500 indoor and 500 outdoor pairs for dehazing, Rain1400 testing set that contains 1,400 pairs for deraining, and Snow100K-L testing set that contains 16,801 pairs for desnowing. For real-world testing sets, we select WeatherStream~\cite{zhang2023weatherstream} testing set that contains 4,500 pairs for dehazing, 4,800 pairs for deraining, and 3,960 pairs for desnowing.

\subsubsection{Implementation Details.}
We train our method on the Pytorch platform, utilizing the Adam optimizer with an initial learning rate of $1 \times 10^{-4}$ that is progressively decreased to $1 \times 10^{-7}$ through a cosine annealing strategy. In addition, we adopt data augmentation, including random cropping, flipping, and rotation, where we randomly crop images to the size of $224 \times 224$. We train TANet for 500k iterations with a batch size of 16. We train and test TANet using an NVIDIA RTX A5000 GPU.     

\subsection{Experimental Results}
In this section, we qualitatively and quantitatively compare TANet with four state-of-the-art all-in-one-image restoration methods, including Transweather~\cite{Valanarasu_2022_CVPR}, KCKE~\cite{Chen2022MultiWeatherRemoval}, WGWS~\cite{zhu2023Weather}, and PromptIR~\cite{potlapalli2023promptir}, and four state-of-the-art multiple degradation image restoration methods, including NAFNet~\cite{chen2022simple}, FocalNet~\cite{cui2023focal}, GRL~\cite{li2023grl}, and MPRNet~\cite{Zamir2021MPRNet}. Note that we train all methods on the mixed dataset as described in Section~\ref{Datasets} and evaluate them separately in different weather conditions.   

\subsubsection{Quantitative Comparisons.}
In Table~\ref{Tab:Test_Synthetic}, we compare TANet with state-of-the-art methods on synthetic datasets, including SOTS for dehazing, Rain1400 for deraining, and Snow100K-L for desnowing. TANet outperforms the state-of-the-art method MPRNet by 1.26dB on average. Especially, TANet outperforms MPRNet by 3.01dB in dehazing, 0.15dB in deraining, 0.62dB in desnowing. Compared to the state-of-the-art all-in-one image restoration method PromptIR, TANet outperforms it by 2.18dB on average. In Table~\ref{Tab:Test_WeatherStream}, we compare TANet with state-of-the-art methods on the real-world WeatherStream dataset, providing real hazy, rainy, snowy images with the corresponding clean images. TANet also achieves state-of-the-art results and outperforms the second-best PromptIR by 0.12dB. Besides, previous all-in-one image restoration methods often contain a large number of parameters for addressing multiple unknown weather conditions, such as 38M in Transweather, 29M in KCKE, and 33M in PromptIR. In contrast, since TANet leverages the inductive bias of adverse weather conditions, TANet can only utilize 9M parameters to achieve state-of-the-art results for all-in-one adverse weather image restoration. Moreover, TANet also runs efficiently with an inference time of 18ms on an NVIDIA RTX A5000 GPU, where we measure the inference time using images of size $256\times 256$.      

\begin{figure}[t!]
\begin{center}
\includegraphics[width=1\columnwidth]{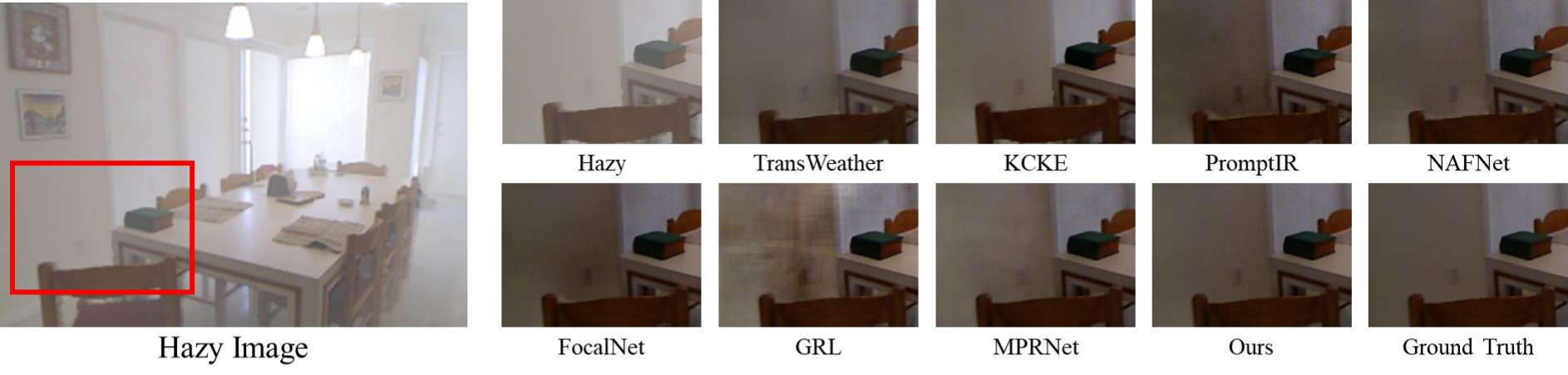}
\end{center}
\caption{Qualitative comparison of dehazing performances on the SOTS~\cite{Resides} test set.}
\label{fig:01_syn_haze}
\end{figure}

\begin{figure}[t!]
\begin{center}
\includegraphics[width=1\columnwidth]{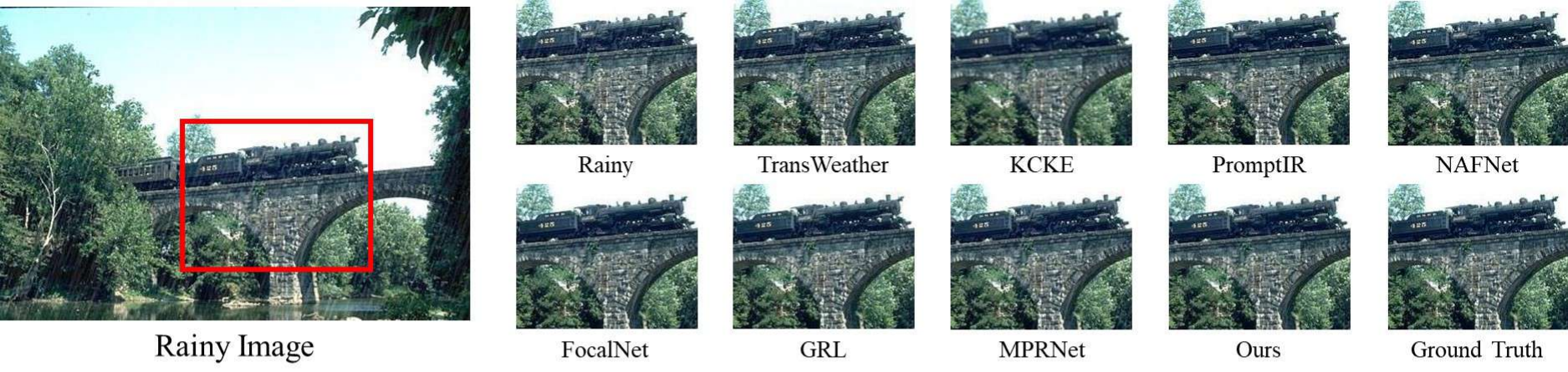}
\end{center}
\caption{Qualitative comparison of deraining results on the Rain1400~\cite{8099669} test set.}
\label{fig:01_syn_rain}
\end{figure}

\begin{figure}[t!]
\begin{center}
\includegraphics[width=1\columnwidth]{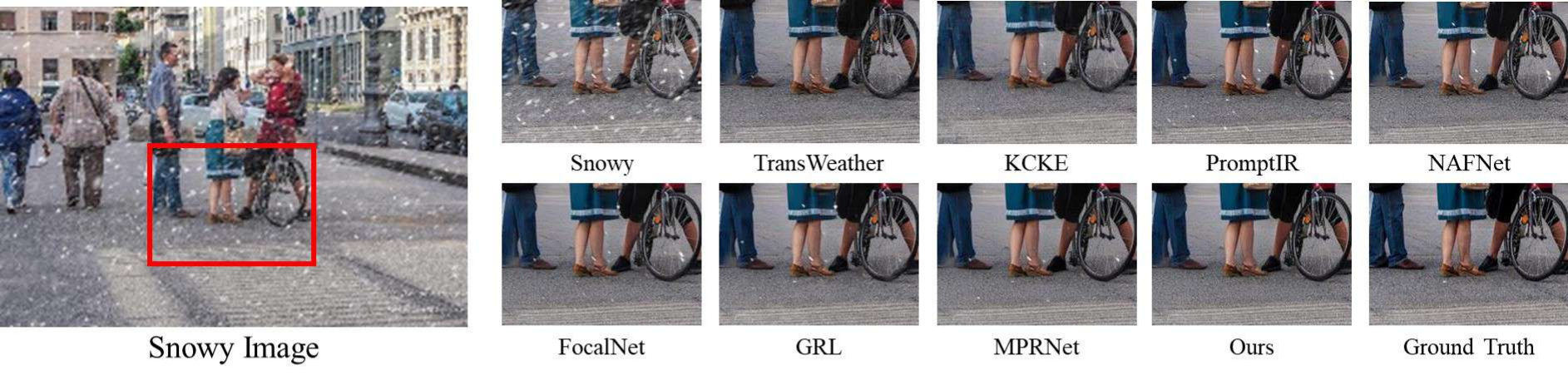}
\end{center}
\caption{Qualitative comparison of desnowing performances on the Snow100K-L~\cite{8291596} test set.}
\label{fig:01_syn_snow}
\end{figure}

\begin{figure}[t!]
\begin{center}
\includegraphics[width=1\columnwidth]{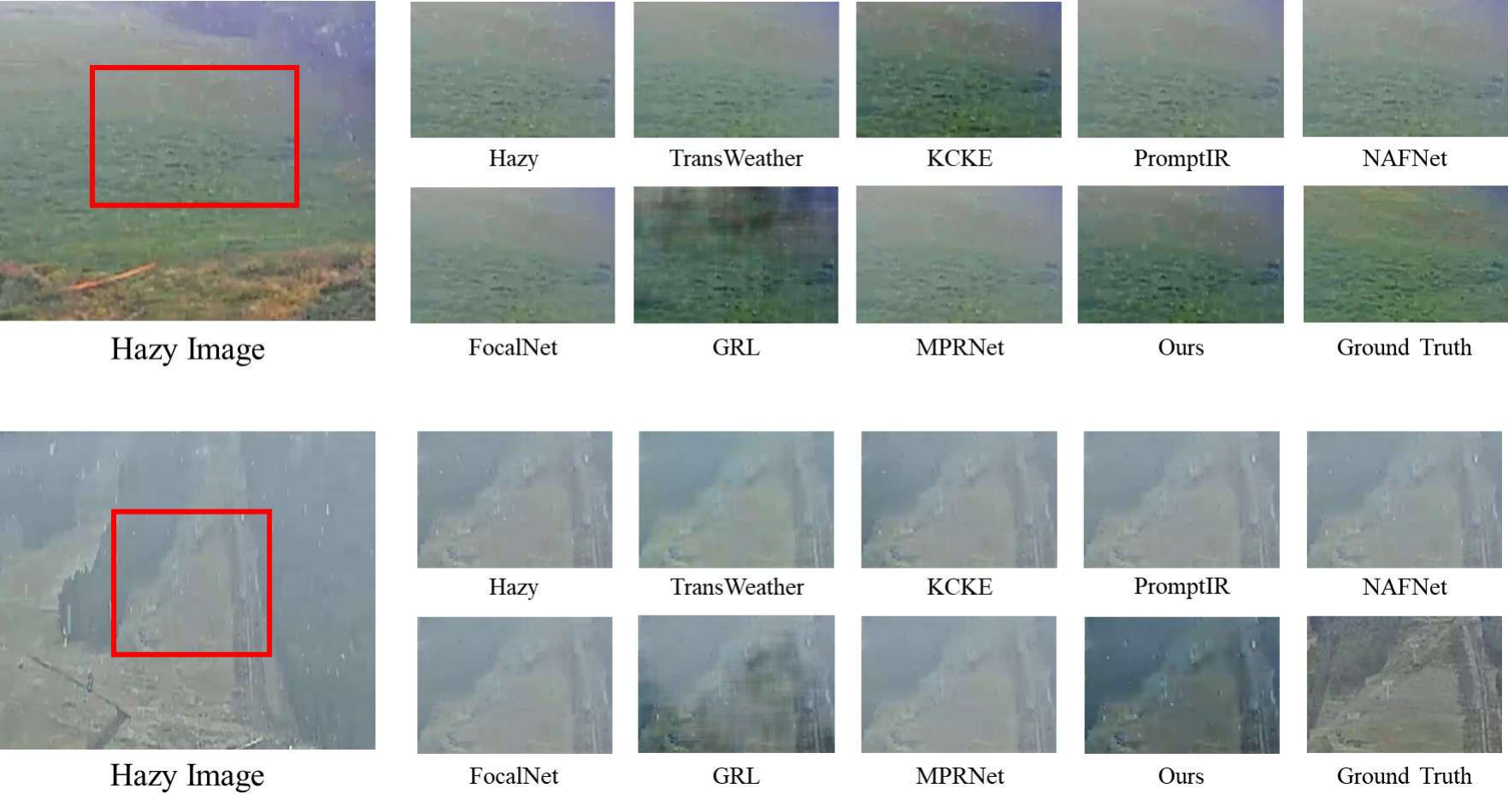}
\end{center}
\caption{Qualitative comparison of dehazing performances on the WeatherStream~\cite{zhang2023weatherstream} dehazing test set.}
\label{fig:03_ws_haze}
\end{figure}

\begin{figure}[t!]
\begin{center}
\includegraphics[width=1\columnwidth]{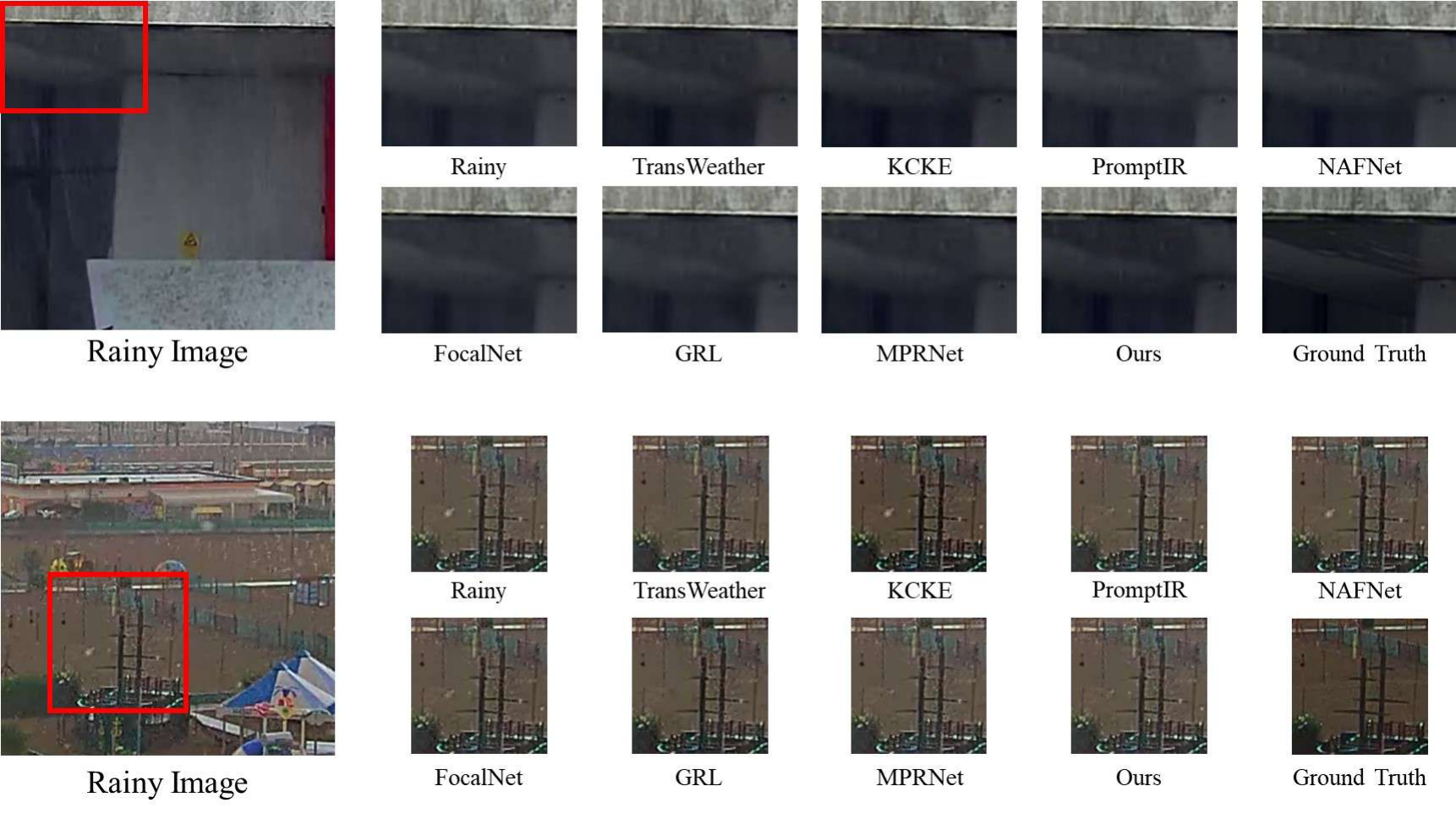}
\end{center}
\caption{Qualitative comparison of deraining performances on the WeatherStream~\cite{zhang2023weatherstream} deraining test set.}
\label{fig:03_ws_rain}
\end{figure}

\begin{figure}[t!]
\begin{center}
\includegraphics[width=1\columnwidth]{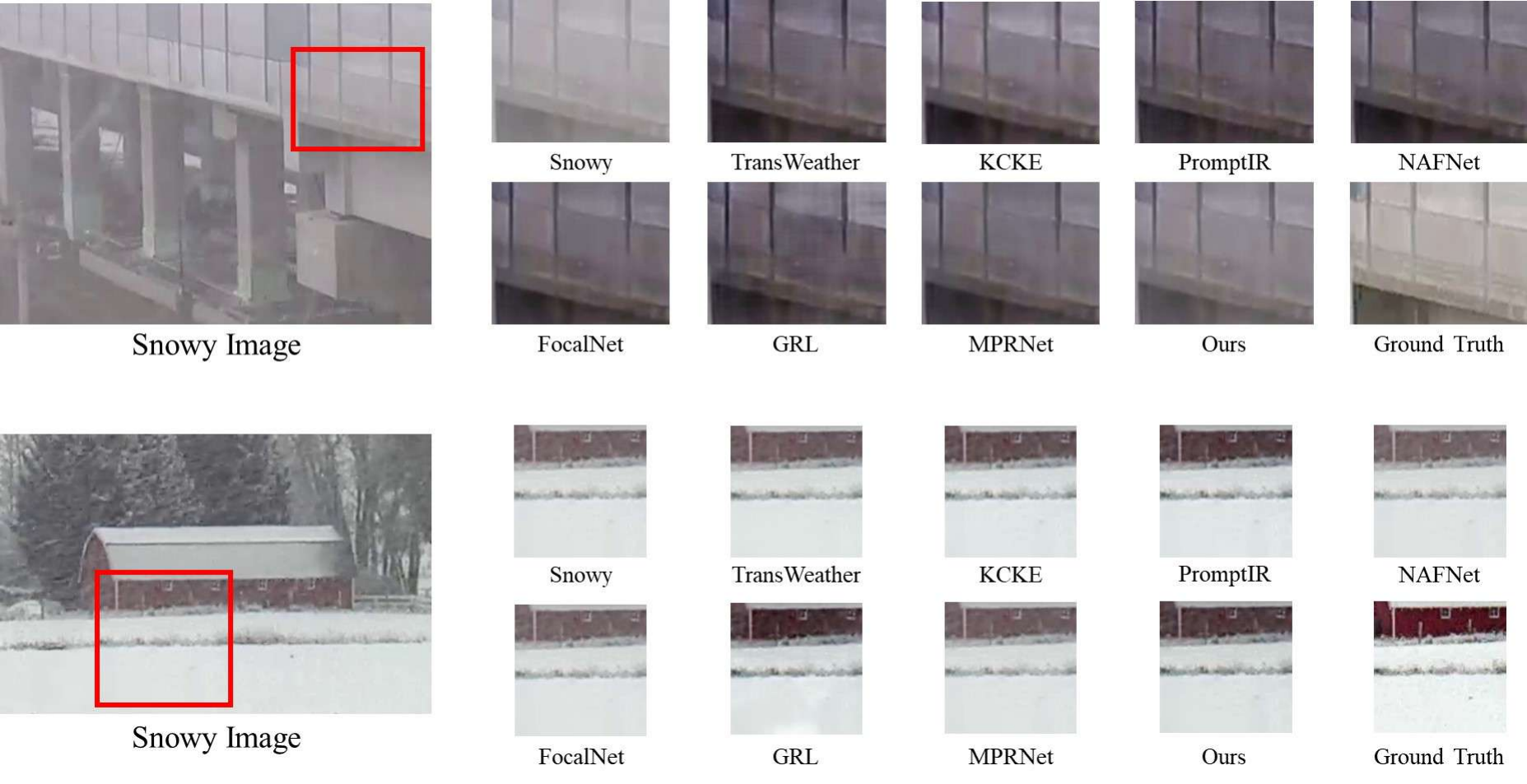}
\end{center}
\caption{Qualitative comparison of desnowing performances on the WeatherStream~\cite{zhang2023weatherstream} desnowing test set.}
\label{fig:03_ws_snow}
\end{figure}

\subsubsection{Qualitative Comparisons.}
In Figures~\ref{fig:01_syn_haze} to ~\ref{fig:01_syn_snow}, we demonstrate the qualitative comparisons on the synthetic datasets, including SOTS for dehazing, Rain1400 for draining, and Snow100K-L for desnowing, respectively. The results show that TANet achieves better or comparable results compared to the previous state-of-the-art methods. In Figures~\ref{fig:03_ws_haze} to ~\ref{fig:03_ws_snow}, we demonstrate the qualitative comparisons on the real-world WeatherStream dataset, including dehazing, deraining, and desnowing results, respectively. TANet also achieves better or comparable results compared to the previous state-of-the-art methods. Particularly, real-world scenarios often involve mixed adverse weather conditions, such as rainy with hazy or snowy with hazy patterns. TANet can still successfully remove such mixed degradation patterns, demonstrating the effectiveness of an all-in-one training strategy.  

\subsection{Ablation Studies}

In this section, we conduct component analyses of the proposed Triplet Attention Block (TAB). TAB consists of three types of attention modules, including Local Pixel-wise Attention (LPA), Global Strip-wise Attention (GSA), and Global Distribution Attention (GDA). We conduct the ablation studies on synthetic datasets, including SOTS testing set for dehazing, Rain1400 testing set for deraining, Snow100K-L testing set for desnowing.  

In Table~\ref{Tab:ablation}, in the first row (Net1), we build a baseline by replacing the proposed components, including LPA, GSA, and GDA, with CNN layers. In the second row (Net2), we demonstrate the effectiveness of LPA, which aims to capture local occlusion artifacts caused by degraded patterns. By using LPA, we can obtain a 0.22dB improvement on average compared to the Net1. In the third row (Net3), we demonstrate the effectiveness of GSA, which aims to extract global occlusion artifacts caused by degraded patterns. By using GSA, we can significantly improve the performance by 1.15dB compared to the Net2. We attribute the success of using GSA in TAB as follows. Since occlusion artifacts under adverse conditions are non-uniform with various orientations and magnitudes, GSA disentangles features into horizontal and vertical directions and resembles them to effectively address degradation patterns with various orientations. Besides, by combining LPA and GSA, TAB can leverage multi-scale features to address non-uniform degraded patterns with various magnitudes. In the fourth row (NET4), we demonstrate the effectiveness of GDA, which aims to capture the distribution of atmospheric particles. GDA can further improve the performance by 0.56dB compared to the Net3. As images under adverse weather conditions often suffer from scattering of atmospheric particles, TANet utilizes GDA to successfully capture the distribution of atmospheric particles, effectively enhancing the quality of restored results. Finally, we demonstrate the effectiveness of using FFT loss in the last row (Net5). Following~\cite{cui2023focal}, we utilize the FFT loss to supervise restoring image in the frequency domain, further improving the performance by 0.36dB compared to the Net4. The ablation studies show that TAB successfully leverages the inductive bias of adverse weather conditions to address degraded patterns by considering the concept of occlusion and scattering.     

\begin{table*}[t!]
\small
\centering
\setlength{\tabcolsep}{3mm}
\caption{Component analysis of Triplet Attention Block (TAB) that consists of Local Pixel-wise Attention (LPA), Global Strip-wise Attention (GSA), and Global Distribution Attention (GDA). FFT denotes the FFT loss.}
\begin{tabular}{l|cccc|cccc}
\noalign{\hrule height 1.0pt}
& LPA & GSA & GDA & FFT & Haze & Rain & Snow  & Average \\
\noalign{\hrule height 1.0pt}
Net1 & & & & & 29.76 & 30.47 & 27.64 & 29.29 \\ 
Net2 & $\surd$ & & & & 30.40 & 30.41 & 27.72 & 29.51 \\ 
Net3 & $\surd$ & $\surd$ & & & 32.36 & 31.02 & 28.61 & 30.66\\ 
Net4 & $\surd$ & $\surd$ & $\surd$ & & 32.78 & 31.38 & 29.50 & 31.22 \\ 
Net5 & $\surd$ & $\surd$ & $\surd$ & $\surd$ & \bf 33.33 & \bf 31.50 & \bf 29.91 & \bf 31.58\\ 
\noalign{\hrule height 1.0pt}
\end{tabular}
\label{Tab:ablation}
\end{table*}

\section{Conclusion}
In this paper, we propose a novel network, called TANet, for all-in-one adverse weather image restoration network. Since images taken under adverse weather conditions often suffer from occlusions, color distortion, and contrast attenuation caused by degraded patterns and the scattering of atmospheric particles, TANet leveraging these common characteristics across multiple weather conditions to effectively restore degraded images in an all-in-one manner. Particularly, TANet utilize Triplet Attention Block (TAB) that contains  Local Pixel-wise Attention (LPA), Global Strip-wise Attention (GSA), and Global Distribution Attention (GDA) to effectively address occlusions and scattering artifacts under adverse weather conditions. By leverage the inductive bias of adverse weather conditions TANet efficiently and effectively achieves state-of-the-art performance in all-in-one adverse weather image restoration.

\begin{credits}
\subsubsection{\ackname} This work was supported in part by the National Science and Technology Council
(NSTC) under grants 112-2221-EA49-090-MY3, 112-2634-F002-005, 111-2221-E-007-046-MY3, and 112-2221-E-007-077-MY3. This work was funded in part by Qualcomm
through a Taiwan University Research Collaboration Project.

\end{credits}
%
%
%
%

\bibliographystyle{splncs04}
\bibliography{main}

\end{document}